%% file: main.tex
\documentclass[10pt,onecolumn,letterpaper]{article}

\usepackage{wacv}
\usepackage{times}
\usepackage{epsfig}
\usepackage{graphicx}
\usepackage{amsmath}
\usepackage{amssymb}
\usepackage{breqn}
\usepackage{booktabs}

\usepackage[pagebackref=true,breaklinks=true,letterpaper=true,colorlinks,bookmarks=false]{hyperref}

\wacvfinalcopy 

\def\wacvPaperID{243} 

\newtheorem{theorem}{Theorem}[section]

\newtheorem{proposition}[theorem]{Proposition}

\def\ie{\textit{i.e.}~}
\def\Vec#1{{\boldsymbol{#1}}}
\def\Mat#1{{\boldsymbol{#1}}}

\ifwacvfinal\fi
\begin{document}

\title{Discovering Discriminative Cell Attributes for HEp-2 Specimen Image Classification}

\author
{
	Arnold Wiliem{\tiny~}$^{1}$, Peter Hobson{\tiny~}$^{2}$, Brian C. Lovell{\tiny~}$^{1}$\\
	~\\
	$^{1}$~The University of Queensland, School of ITEE, QLD 4072, Australia\\
	$^{2}$~Sullivan Nicolaides Pathology, QLD 4068, Australia\\
	{\tt\small a.wiliem@uq.edu.au, lovell@itee.uq.edu.au, peter_hobson@snp.com.au}
}


\maketitle
 \thispagestyle{empty}

\begin{abstract}

Recently, there has been a growing interest in developing Computer Aided Diagnostic (CAD) systems for improving the reliability and consistency of pathology test results. 
This paper describes a novel CAD system for the Anti-Nuclear Antibody (ANA) test via Indirect Immunofluorescence protocol on Human Epithelial Type 2 (HEp-2) cells.
While prior works have primarily focused on classifying cell images extracted from ANA specimen images, this work takes a further step by focussing on the specimen image classification problem itself.
Our system is able to efficiently classify specimen images as well as producing meaningful descriptions of ANA pattern class which helps physicians to understand the differences between various ANA patterns.
We achieve this goal by designing a specimen-level image descriptor that: (1)~is highly discriminative; (2)~has small descriptor length and (3)~is semantically meaningful at the cell level.
In our work, a specimen image descriptor is represented by its overall cell attribute descriptors.
As such, we propose two max-margin based learning schemes to discover cell attributes whilst still maintaining the discrimination of the specimen image descriptor.
Our learning schemes differ from the existing discriminative attribute learning approaches as they primarily focus on discovering image-level attributes.
Comparative evaluations were undertaken to contrast the proposed approach to various state-of-the-art approaches on a novel HEp-2 cell dataset which was specifically proposed for the specimen-level classification.
Finally, we showcase the ability of the proposed approach to provide textual descriptions to explain ANA patterns.

\end{abstract}

\input{sec_introduction}


\input{sec_problemdefinition}
\input{sec_proposedapproach}

\input{sec_experiment}

\input{sec_conclusion}
\input{sec_acknowledgement}
\vspace{-1ex}
\input{sec_appendix}

{\small
\bibliographystyle{ieee}
\bibliography{egbib}
}

\end{document}

%% file: sec_introduction.tex
\vspace{-2ex}
\section{Introduction}

\begin{figure}[!tb] 
\centering
\includegraphics[width=0.5\columnwidth]{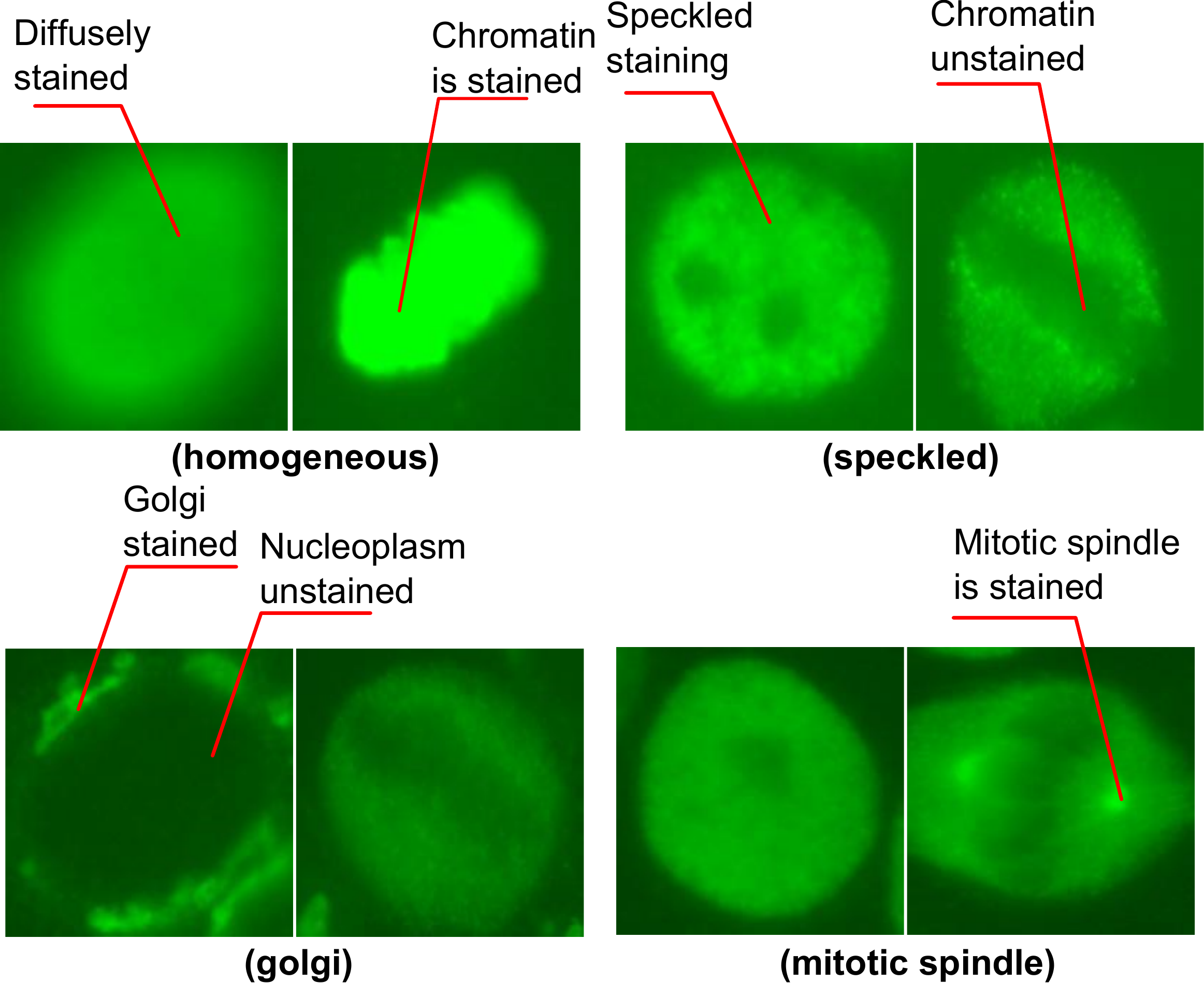}
  \caption {Examples of {cell-level} discriminative attributes found by our proposed approach for some ANA patterns. 
  Each pattern is described using two cell types: interphase cell (left); mitotic cell (right).}
  \label{fig:examples}
\end{figure}

The application of image analysis for various routine clinical pathology tests has been growing in recent years~\cite{foggia2013benchmarking,
labati2011lymphoblastic}. 
When incorporated into subjective analysis from scientists, these can potentially not only lower test turn around time but also increase test result reliability and consistency across laboratories~\cite{foggia2013benchmarking,Wiliem2011}.

The common way of identifying the existence of connective tissue diseases is via the Anti-Nuclear Antibody (ANA) test using Indirect Immunofluorescence (IIF) protocol on Human Epithelial type 2 (HEp-2) cells~\cite{meroni2010ana}. 
This is due to its high sensitivity and the large range expression of antigens. 
Unfortunately, the protocol is time consuming, labour intensive and subjective~\cite{Bizzaro1998,Pham2005} leading to low reproducibility and large inter/intra- personnel/laboratory variations~\cite{Soda2009}.
One possible solution is to apply Computer Aided Diagnostic (CAD) systems for automated classification of ANA IIF digitally captured images.

\begin{figure}[!tb] 
\includegraphics[width=\columnwidth]{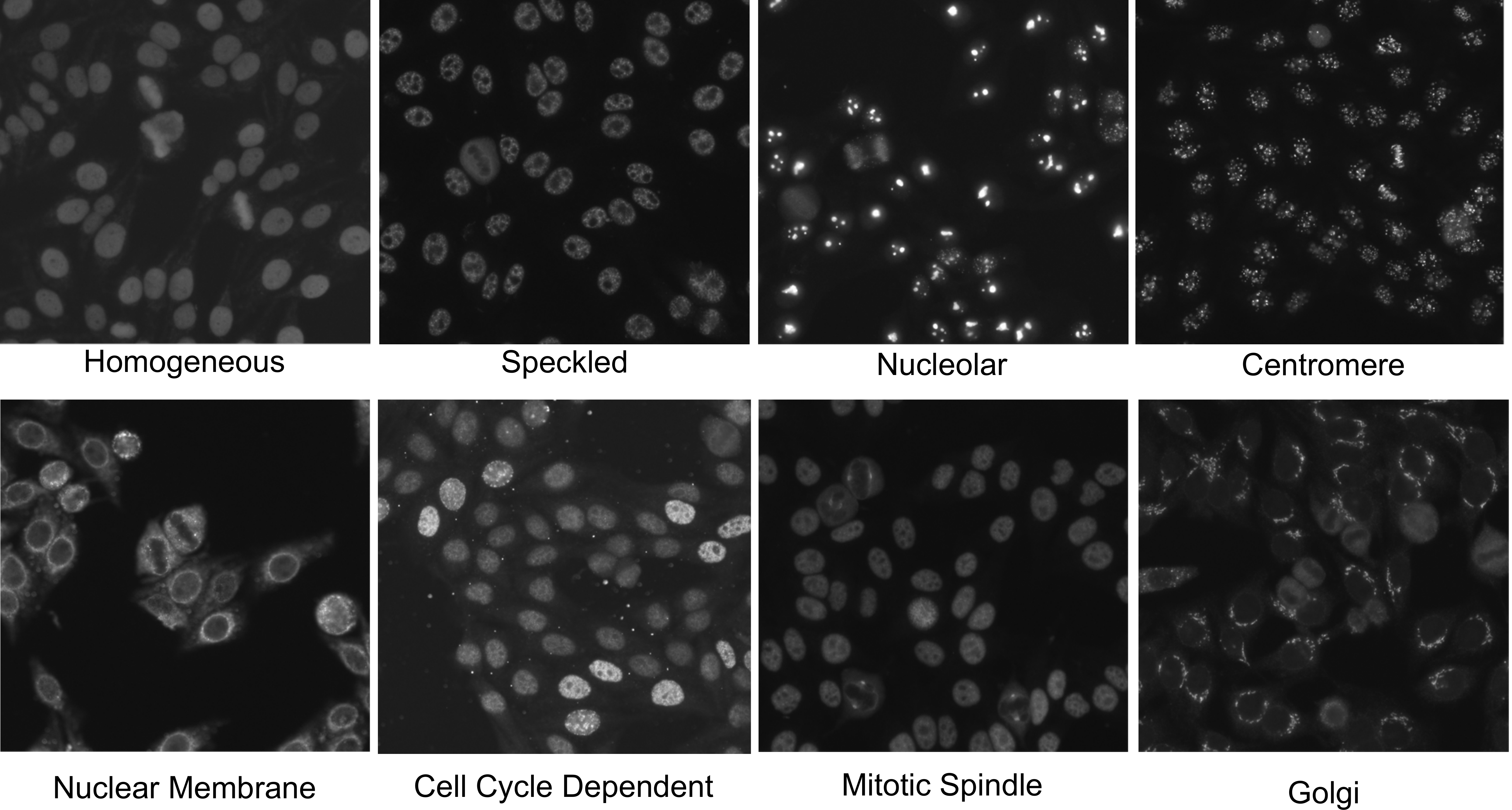}
  \caption {Sample images from the proposed dataset}
  \label{fig:dataset}
\end{figure}

Despite the large interest shown recently in the literature, most of the existing works only focus on the early steps of the CAD system, that is {cell-level} classification~\cite{foggia2013benchmarking,Wiliem2011,Wiliem2013_wacv,Wiliem2013,Faraki2013,Yang2013}. 
Whilst, some methods which go beyond this scope assume that the {specimen-level} pattern can be simply estimated from the most dominant {cell-level} pattern~\cite{Soda2009}.
As each {specimen-level} image consists of a set of cells, ideally a CAD system should be able to extract more useful information from the cells distribution to infer the specimen image pattern (refer to Fig.~\ref{fig:dataset} for examples of specimen images).
Furthermore, the existing systems do not provide meaningful information as to why an ANA pattern differs to the others.
Having meaningful information such as textual description is of interest in this area since often a pattern has various descriptions amongst physicians~\cite{wiik2010ana}. 

One way to address these issues is to learn discriminative and semantically meaningful descriptors. 
Each element of the descriptor defines the existence/absence of a specific inherent property/characteristics in an image.
For instance, an image containing a car may have some properties such as \textit{has a wheel}, \textit{is metallic}, \textit{has windows and doors}~\cite{Ferrari07}.
These properties are popularly known as image attributes~\cite{Ferrari07}.

There is a growing interest to develop attribute-based approach for image classification~\cite{Ferrari07,
LampertCVPR2009,
Bergamo2011,
RastegariECCV2012,
parikh2009}. 
For instance, Ferrari and Zisserman proposed a probabilistic generative model of visual attributes~\cite{Ferrari07}. 
Lampert~\etal reported excellent results of the Direct Attribute Prediction (DAP) approach in a \textit{zero-shot} learning problem~\cite{LampertCVPR2009}.
Parikh~\etal extended the notion of image attribute into relative attribute which is related to adjective of a noun such as \textit{larger} and \textit{more open space}~\cite{parikh2009}.

An image attribute detector is essentially a binary classifier which determines the presence or absence of an image property.
As such, each attribute needs a training set which may be expensive to acquire for our problem domain due to the limited number of domain experts.
It is also almost impossible to use the \textit{Amazon Mechanical Turk} service~\cite{parikh2009,LampertCVPR2009} to acquire the training labels.
Therefore, it is desirable to automatically discover the smallest set of discriminative image attributes wherein the domain experts can name them.
To that end, one could apply ideas proposed in~\cite{Bergamo2011,RastegariECCV2012} to discover discriminative image attributes.
Here, the attribute classifiers are jointly learned with the image classifier in the max-margin framework.
Nevertheless, these approaches are mainly focussed on image attributes, whereas in our work, the system needs to discover {cell-level} attributes which, when summarised into the {specimen-level} descriptor, will be highly discriminative.
Henceforth, the cell attributes must be indirectly discovered via learning a discriminative specimen image descriptor.

The concept of discovering discriminative image attributes is also related to hashing techniques~\cite{charikar2002similarity,kulis2009kernelized,gong2012,weiss2008nips}.
However, the descriptors resulting from these techniques are not necessarily semantically meaningful.
For instance the spectral hashing approach aims to generate a set of binary codes which are short, easy to compute and maps similar items to similar binary codewords~\cite{weiss2008nips}. 
To that end, the approach constraints that each bit has a 50\% chance of being zero and one and different bits independent to each other.
These are much weaker constraints to discover semantically meaningful code.

\textbf{Contributions} The aim of the present work is to devise an algorithm which learns image descriptors for ANA IIF specimen image classification problem with three properties: (1)~highly discriminative; (2)~semantically meaningful at cell level and (3)~having short descriptor length. 
We achieve this by proposing two learning schemes for discovering {cell-level} attributes through a discriminative learning framework.
In contrast to previous approaches~\cite{Bergamo2011,RastegariECCV2012} which learns discriminative {image-level} attributes, our approach learns {cell-level} attributes where their values can be used to construct discriminative {image-level} descriptors.
Our theoretical results show that under a certain condition, it is possible to devise solutions based on {image-level} discriminative attribute learning for solving the posed problem.
Finally, we further showcase (refer to Fig.~\ref{fig:examples} for some examples of discovered meaningful cell attributes) that a textual description can be generated from the learned {cell-level} attributes and shares similarities to the description from experts.
We evaluated all the approaches on the new HEp-2 cell dataset proposed for the specimen image classification problem.
To our knowledge this is the first comprehensive dataset constructed for this purpose.

We continue our discussion as follows. 
Section~\ref{sec:problem_def} discusses the ANA IIF {specimen-level} classification problem.
The proposed learning schemes are described in Section~\ref{sec:proposed_approach}. 
We present the experiment and results in Section~\ref{sec:experiment}.
Finally, the main findings and future direction are discussed in Section~\ref{sec:conclusions}.

%% file: sec_problemdefinition.tex
\section{Problem definition}
\label{sec:problem_def}

An ANA IIF specimen image $\Mat{I}$ is represented by the three-tuple $\{\Mat{I}, \Mat{M}, \delta\}$ which consists of: \textbf{(i)} the Fluorescein Isothiocyanate (FITC) image channel which carries pattern information $\Mat{I}$; \textbf{(ii)} a binary cell mask image $\Mat{M}$ which are extracted from the \textit{4',6-diamidino-2-phenylindole} (DAPI) image channel; \textbf{(iii)} the fluorescence intensity $\delta = \{\operatorname{weak}, \operatorname{strong} \}$. 
The goal is to construct a classifier which classifies a specimen image into one of the known classes.

Our problem differs from~\cite{foggia2013benchmarking,Wiliem2013_wacv,Wiliem2013,
Faraki2013,Yang2013} in the way that these works focus on classification of individual cell images extracted from specimen images; our main attention is on the specimen image classification problem.

%% file: sec_proposedapproach.tex
\section{Discovering cell attributes}
\label{sec:proposed_approach}


The goal of the present work is to discover {cell-level} attributes which can be used to form a discriminative {specimen-level} descriptor.
Once all cells are extracted from a specimen image using its mask, each cell image is divided into J regions from which a regional descriptor is extracted. 
For clarity, we defer the discussion of how the regions are divided until Section~\ref{sec:experiment_settings}.
After the cell has been divided into regions, we derive the {cell-level} attributes from the extracted regional descriptor.
The specimen image descriptor is then formed by concatenating the overall cell attributes from all regions.
Let $\Vec{z}_i \in \mathbb{R}^P$ be the $P$ dimensional {specimen-level} descriptor of the $i$-th specimen image, $\Vec{z}_i = [\hat{\Vec{h}}_{i,1}  \dots \hat{\Vec{h}}_{i,J}]$.
Each $\hat{\Vec{h}}_{i,j} \in \mathbb{R}^b$ represents the overall {cell-level} attribute descriptor extracted from the $j$-th region:

\vspace{-1ex}
\begin{equation}
	\label{eqn:average_h_j}
	\hat{\Vec{h}}_{i,j} = \frac{1}{N_{i,j}} \sum^{N_{i,j}}_{c=1}{\Vec{h}_{i,j,c}}
\end{equation}

\vspace{-1ex}
\noindent
where $N_{i,j}$ is the number of cells extracted from the specimen image for $j$-th region; $\Vec{h}_{i,j,c} \in \mathbb{R}^b$ is the {cell-level} attribute descriptor extracted from the $j$-th region in the $c$-th cell; $b$ is the number of cell attributes extracted from $j$-th region.

The above equation suggests that the specimen image is represented by the average of the {cell-level} attributes. 
This approach differs from~\cite{Soda2009,foggia2013benchmarking} wherein each image is represented as the dominant pattern of extracted cells.
We will later show in the experiment that our strategy is considerably more effective.

Inspired from~\cite{Bergamo2011}, we define the value of each element of the {cell-level} attribute descriptor $\Vec{h}_{i,j,c}$ as the output of a set of basis linear classifiers as follows:

\vspace{-2ex}
\begin{equation}
	\label{eqn:h_j}
	\Vec{h}_{i,j,c} = \Mat{A}_j^{\top}\Vec{x}_{i,j,c}
\end{equation}

\vspace{-1ex}
\noindent
where each column of $\Mat{A_j} \in \mathbb{R}^{d \times b}$ is the model parameter for a single basis classifier; $\Vec{x}_{i,j,c} \in \mathbb{R}^d$ is the regional descriptor extracted from $j$-th region in $c$-th cell.


The value of each element in $\Vec{h}_{i,j,c}$ indicates the presence ($+$) or absence ($-$) of a particular {cell-level} attribute. 
Here, each attribute classifier will be trained by maximising the margin between the positive and negative samples, 
thereby, imposing an indirect constraint that an attribute should provide a meaningful concept (\ie the concept discriminating positive and negative samples).

We learn both the {image-level} classifier and the {cell-level} attribute basis classifiers simultaneously over all variables.
To this end, we propose two learning schemes via the one-versus-all linear SVM framework. 
The first scheme, denoted \textit{All Region Cell Attribute Descriptor} (ARCAD), is designed to discover discriminative {cell-level} attributes extracted from each region at the same time.
On the other hand, the second scheme, namely \textit{Cell Regional Attribute Descriptor} (CRAD), is proposed to discover the most discriminative {cell-level} attributes for each individual region.

Let $\mathcal{G} \in \{(\Mat{I}_i,\Vec{y}_i)\}^N_{i=1}$ be the training sets wherein each image $I_i$ has label vector $\Vec{y}_i \in \{-1,+1\} ^ K$ encoding the class label from K possible patterns. 
For instance, if $I_i$ belongs to the first class, then $\Vec{y}_{i,1} = +1$ and $-1$ for the rest of its elements.
Let $\mathcal{X} \in \{ \{\Vec{x}_{i,j,1}\}_{j=1}^{J} \dots  \{\Vec{x}_{i,j,N_{i,j}}\}_{j=1}^{J}\}$ be the set of extracted regional descriptors of the specimen image $I_i$.

\subsection{All Region Cell Attribute Descriptor (ARCAD)}

The training objective for ARCAD is defined by:

\begin{center}
\vspace{-2ex}
\begin{small}
\begin{equation}
\min_{\Mat{w}_{1 \dots K}, \Vec{b}_{1 \dots K}} 
	\sum_{k=1}^K \left\{ \frac{1}{2} \| \Mat{w}_k \|^2 + \frac{\lambda}{N} \sum_{i=1}^N \ell \left[ y_{i,k} (\Vec{b}_k + \Mat{w}^{\top}_k \Vec{z}_i \right] \right\}
\end{equation}
\end{small}
\end{center}

\vspace{-2ex}
\noindent
where $\Vec{w}_k$, $\Vec{b}_k$ and $\lambda$ are the hyperplane, bias term and regularisation parameters for each SVM, respectively; $\ell [ \cdot ]$ is the hinge loss function.
If we expand the $\Vec{z}_i$ in the above equation by substituting Eqn.~\ref{eqn:average_h_j} into~\ref{eqn:h_j}, it becomes:

\begin{center}
\vspace{-3ex}
\begin{small}
\begin{equation}
\label{eqn:ARCAD}
\min_{\Mat{w}_{1 \dots K}, \Vec{b}_{1 \dots K}, \Mat{A}_{1 \dots J}} 
	\sum_{k=1}^K \left\{ \frac{1}{2} \| \Mat{w}_k \|^2 + \frac{\lambda}{N} \sum_{i=1}^N \ell \left[ y_{i,k} (\Vec{b}_k +
\sum^{J}_{j=1} \left( \frac{\Vec{w}_{k,j}}{N_{i,j}} \sum^{N_{i,j}}_{c=1} \Mat{A}^{\top}_j \Vec{x}_{i,j,c}  \right) \right] \right\}
\end{equation}
\end{small}
\end{center}

\vspace{-1ex}
\noindent
where $\Vec{w}_{k,j}$ is the parameters for $k$-th 
linear SVM of the $j$-th 
region~\footnote{Here $\Vec{w}_{k,j} \in \mathbb{R}^{1 \times b}$ is a sub-vector of $\Vec{w}_{k} = [\Vec{w}_{k,1} \dots \Vec{w}_{k,J}]$.}.
Minimising the above equation requires to learn all the parameters simultaneously. 
This includes the specimen image one-versus-all SVMs as well as the {cell-level} attribute basis classifiers $\{\Mat{A}_j\}^J_{j=1}$. 

We note that although the Eqn.~\ref{eqn:ARCAD} shares similarities to the objective function discussed in the PiCoDes 
approach of~\cite{Bergamo2011}, applying the solution proposed in PiCoDes to solve the above equation is not straightforward.
This is because in contrast to PiCoDes, the above equation is not aimed at learning discriminative {image-level} attributes.
In the present work, the {image-level} descriptor is formed by concatenating the overall {cell-level} attribute descriptors extracted from cell regions.


For completeness, we present the objective function of PiCoDes which discovers discriminative {image-level} attributes:
\begin{center}
\begin{small}
\begin{equation}
\label{eqn:PiCoDes}
\min_{\Mat{w}_{1 \dots K}, \Vec{b}_{1 \dots K}, \Mat{A}_{1 \dots J}} 
	\sum_{k=1}^K \left\{ \frac{1}{2} \| \Mat{w}_k \|^2 + \frac{\lambda}{N} \sum_{i=1}^N \ell \left[ y_{i,k} (\Vec{b}_k +
\sum^{P}_{p=1} \left(  w_{k,p} \Mat{A}^{\top}_p \Vec{u}_i  \right) \right] \right\}
\end{equation}
\end{small}
\end{center}

\noindent
where $P$ is the number of {image-level} attributes; $w_{k,p}$ is the $p$-th parameter value of the $k$-th SVM model~\footnote{Note that $w_{k,p}$ is a scalar value. $w_{k,p}$ differs from  $\Vec{w}_{k,j}$ which is a sub-vector}; $\Vec{u}_i$ is the $i$-th {image-level} descriptor. Note that $\Vec{u}_i$ is not the same as $\Vec{z}_i$ as the former is image descriptor extracted using various image features.

To address our problem, we present the following proposition.

\vspace{-1ex}
\begin{proposition}
\label{prop:1}
	The problem presented in Eqn.~\ref{eqn:ARCAD} is equivalent to Eqn.~\ref{eqn:PiCoDes} if and only if $\Vec{z}_i$ is formed by $\hat{\Vec{h}}_{i,j}$ defined in Eqn.~\ref{eqn:average_h_j} and $\Vec{u}_i = [\hat{\Vec{x}}_{i,j} \dots \hat{\Vec{x}}_{i,J}]$ where $\hat{\Vec{x}}_{i,j} =  \frac{1}{N_{i,j}} \sum^{N_{i,j}}_{c=1} \Vec{x}_{i,j,c}$
\end{proposition} 

\vspace{-1ex}
The proof for Proposition.~\ref{prop:1} is presented in the Appendix.
The above proposition allows us to design a 
tractable solution for Eqn.~\ref{eqn:ARCAD} 
using the existing PiCoDeS solution for Eqn.~\ref{eqn:PiCoDes}.


\subsection{Cell Regional Attribute Descriptor (CRAD)}

The ARCAD objective function learns {cell-level} attributes for all regions at the same time.
Another alternative is to learn discriminative {cell-level} attribute from each region exclusively.
To this end, we propose an objective function consisting of 
the summation of $\operatorname{E} ( \cdot )$, defined by:

\begin{small}
\begin{equation}
\operatorname{min} \sum_{j = 1}^J \operatorname{E} \left( \Vec{w}_{1 \dots K}^{[j]}; \Vec{b}^{[j]}_{1 \dots K}; \Mat{A}_j; \{ ( \hat{\Vec{h}}_{i,j}, \Vec{y}_i ) \}_{i = 1}^N \right) 
\end{equation}
\end{small}

\noindent
where $\operatorname{E} ( \cdot )$ is the objective function for learning the basis classifier for each {cell-level} attribute extracted from each region:
\begin{small}
\begin{equation}
	\operatorname{E} \left( \Mat{A}_j; \{ \hat{\Vec{h}}_{n,j} \}_{n = 1}^N \right) =  
	\sum_{k=1}^K \left\{ \frac{1}{2} \| \Mat{w}^{[j]}_k \|^2 + \frac{\lambda^{[j]}}{N} \sum_{i=1}^N \ell \left[ y_{i,k} (\Vec{b}^{[j]}_k + \Mat{w}^{[j]\top}_k \hat{\Vec{h}}_{i,j} \right] \right\}
\end{equation}
\end{small}

\noindent
where $\Vec{w}^{[j]}_k$ and $\Vec{b}^{[j]}_k$ are the $k$-th SVM parameters of the $j$-th region {cell-level} attributes.
Similar to Eqn.~\ref{eqn:ARCAD}, the above equation can 
be expanded by substituting Eqn.~\ref{eqn:average_h_j} into~\ref{eqn:h_j}. We get:
\begin{small}
\begin{equation}
\label{eqn:CRAD}
	\operatorname{E} \left( \Mat{A}_j; \{ \hat{\Vec{h}}_{n,j} \}_{n = 1}^N \right) = \sum_{k=1}^K \left\{ \frac{1}{2} \| \Mat{w}^{[j]}_k \|^2 + \frac{\lambda^{[j]}}{N} \sum_{i=1}^N \ell \left[ y_{i,k} (\Vec{b}^{[j]}_k + \frac{\Mat{w}^{[j]\top}_k}{N_{i,j}} \sum_{c=1}^{N_{i,j}} \Mat{A}^{\top}_j \Vec{x}_{i,j,c} \right] \right\}
\end{equation}
\end{small}

\noindent
This brings us to similar challenge as posed in Eqn.~\ref{eqn:ARCAD} which can be solved by using Proposition.~\ref{prop:1}.
We note that the purpose of Eqn.~\ref{eqn:CRAD} is only to train the cell attribute basis classifiers from each region individually.
Once these are trained, we train the one-versus-all SVM classifiers
for the specimen-level classification.


\subsection{Optimisation algorithm}
\label{sub:optimisation_algorithm}

We present the optimisation algorithm used to solve Eqn.~\ref{eqn:PiCoDes} for solving Eqn.~\ref{eqn:ARCAD} and~\ref{eqn:CRAD}.
The algorithm uses block coordinate descent alternating between optimising the 
SVM parameters and the attribute basis classifiers.

\subsubsection{Learning specimen image SVM parameters}

When $\Mat{A}_j$ is fixed, $\Vec{z}_i$ can be determined.
Thus, the problem is reduced to linear a SVM learning problem for both ARCAD and CRAD.

\subsubsection{Learning attribute basis classifiers}

More elaborate steps are required to learn individual 
attribute basis classifiers. 
It has been shown in~\cite{Bergamo2011} that when $\Vec{w}_k$ and $\Vec{b}_k$ are fixed, updating each {cell-level} attribute base classifier from each region can be done via learning the upper
bound of the objective function which resembles to learning 
a linear SVM classifier by minimising the sum of weighed
mis-classifications. Our objective function is:~\footnote{Readers are encouraged to read~\cite{Bergamo2011} for further details.}

\vspace{-1ex}
\begin{equation}
	 \operatorname{\hat{E}} (\Vec{a_{b}}) = \sum^N_{i=1} v_i  \ell( q_i \Vec{a_b} \hat{\Vec{x}}_{i,j} )
	 \label{eqn:basis_learning}
\end{equation}

\vspace{-1ex}
\noindent 
where $\hat{\Vec{x}}_{i,j}$ is the average of the low level features extracted from the $j$-th region of image $I_i$; $\Vec{a_b}$ is the base classifier model of the \mbox{$b$-th} attribute; $q_i \in \{-1,+1\}$ and $v_i \in \mathbb{R}^{+}$ are known values computed via:

\vspace{-2ex}
\begin{align}
	q_i & = \left| \sum^{K}_{k=1} \ell (\alpha_{i,k,b} + \beta_{i,k,b} ) - \ell( \beta_{i,k,b} ) \right| \\
	v_i & = \operatorname{sgn} \left( \sum^{K}_{k=1} \ell (\alpha_{i,k,b} + \beta_{i,k,b} ) - \ell (\beta_{i,k,b}) \right)
\end{align}

\vspace{-1ex}
\noindent
where $\alpha$ and $\beta$ are defined by: $\alpha_{i,k,b} \equiv y_{i,k} w_{k,b}$ and $\beta_{i,k,b} \equiv y_{i,k} b_k + \sum_{b' \neq b} y_{i,k} w_{k,b'}\Vec{a_b^{\top}} \Vec{u}_i$

The main difference between CRAD and ARCAD is that CRAD uses the SVM parameters of each region (\ie ~$\Vec{w}^{[j]}_k$ and $\Vec{b}^{[j]}_k$) to update the base classifiers, whereas ARCAD uses the overall SVM parameters.
In other words, in ARCAD, the information learned from 
other regions are used to optimise Eqn.~\ref{eqn:basis_learning}, 
while this is not the case for CRAD.

%% file: sec_experiment.tex
\section{Experiment}
\label{sec:experiment}

We first describe the novel {specimen-level} HEp-2 cell image classification dataset and experiment settings.
The proposed approaches are compared to various notable hashing techniques~\cite{charikar2002similarity,kulis2009kernelized,gong2012,weiss2008nips} as well as recent discriminative attribute learning methods~\cite{RastegariECCV2012,Bergamo2011}.
The proposed approaches are also compared to the existing methods for {specimen-level} image classification in this domain such as the Multiple Expert System (MES)~\cite{Soda2009}.
Finally, we showcase the ability of the discovered {cell-level} attributes to help physicians in describing the ANA pattern classes.

\subsection{Dataset}

We propose a new dataset~\footnote{Available at \url{http://www.itee.uq.edu.au/sas/datasets}} 
to serve our goal because the existing benchmarking datasets such as SNPHEp-2~\cite{Wiliem2013_wacv}, 
ICIP2013~\footnote{\url{http://nerone.diiie.unisa.it/contest-icip-2013/index.shtml}} and 
ICPR2012Contest~\cite{Cordelli2011} are primarily focussed on the {cell-level} classification problem.
Although the ICPR2012Contest dataset provides specimen images, 
the number of images for each train and test are insufficient for this study 
(\ie only 14 images for each train and test classified into six classes).

The proposed dataset was obtained from 262 patient sera in 2013.
The patient sera were diluted to 1:80 dilution.
The prepared specimen was then photographed using a monochrome 
high dynamic range cooled microscopy camera fitted on a microscope.
In total there were 262 specimen images classified into eight classes: homogeneous, speckled, nucleolar, centromere, nuclear membrane (NuMem), cell cycle dependent (CCD), mitotic spindle (MitSp) and golgi aparatus (golgi).

Each specimen image contains a collection of interphase and mitotic cells which are important for the specimen-level classification (refer to Fig.~\ref{fig:dataset}).
The mitotic cells are the HEp-2 cells undergoing the mitosis phase. 
In this phase each cell divides itself into two separate individual cells.
The mitotic cells are of importance as they produce a set of antigens which are either less concentrated or undetectable in the interphase stage.
Due to this fact, experts consider the pattern of both interphase and mitotic cells in classifying an ANA specimen.

Five-fold validations of training and test were created by randomly selecting from the pool of images.
Specifically, we randomly selected approximately 130 images from the image pool for each fold. 
Then, the selected images were further divided into two equally sized sets for training and testing (\ie approximately 75 images for each set).
To our knowledge this is the first dataset to offer a 
reasonable number of data samples for benchmarking 
{specimen-level} classification approaches. We plan to release the dataset by the end of 2014.

\subsection{Experiment settings}
\label{sec:experiment_settings}

Although one could use any {cell-level} descriptor, 
in the present work, we use the bag of words descriptor with Cell Pyramid Matching (CPM) structure recently proposed in~\cite{Wiliem2013} due to its robustness to various laboratory settings.
Specifically, we divide each cell into an inner region, 
which covers the cell content and an outer region, which contains information related to cell edges and shape.
In addition, a descriptor extracted from the whole cell region is also used.
Thus, there are three regional descriptors extracted from each cell.
As there are two cell types: interphase and
mitotic cells, we use six regions in total (three regions of each type).
Note that, we only consider the regional descriptors extracted from the same cell type when computing $\hat{\Vec{x}}_{i,j}$ as well as $\hat{\Vec{h}}_{i,j}$.

Once the regional descriptors are extracted, 
we lift up the each descriptor into a higher-dimensional space approximating the histogram intersection kernel by using the explicit feature maps proposed by Vedaldi and Zisserman~\cite{vedaldi11efficient}. 
Specifically we map the original regional descriptor into the space three times larger by setting the parameter $n$ in~\cite{vedaldi11efficient} to 1.
We opt to use the lifted space of the histogram intersection 
kernel as bag of words histogram comparisons are generally best to be done under histogram intersection kernel as suggested in~\cite{Bergamo2011}.
The lifted-up descriptor will be the regional descriptor for each $\Vec{x}_{i,j,c}$.

Although it is possible to vary the number of attributes extracted from each region, we choose to use the same number of attributes for all regions in order to reduce the total number of combinations.

\subsection{Evaluation of different descriptors}

We first contrast the proposed approaches to the recent hashing methods such as: Locality-Sensitive Hashing (LSH)~\cite{charikar2002similarity}, Kernelised LSH (KLSH)~\cite{kulis2009kernelized}, Spectral Hashing (SPH)~\cite{weiss2008nips} and Iterative Quantization (ITQ)~\cite{gong2012}.
In addition we also compare the proposed approaches to the recent discriminative attributes learning: PiCoDeS~\cite{Bergamo2011} and Discriminative Binary Code (DBC)~\cite{RastegariECCV2012}. 

All methods were trained and tested using the {specimen-level} descriptor extracted with the setup previously described.
We used the unlifted regional descriptor in conjunction with the histogram intersection kernel for KLSH.

\begin{figure}[!tb] 
\centering
\includegraphics[width=0.7\columnwidth]{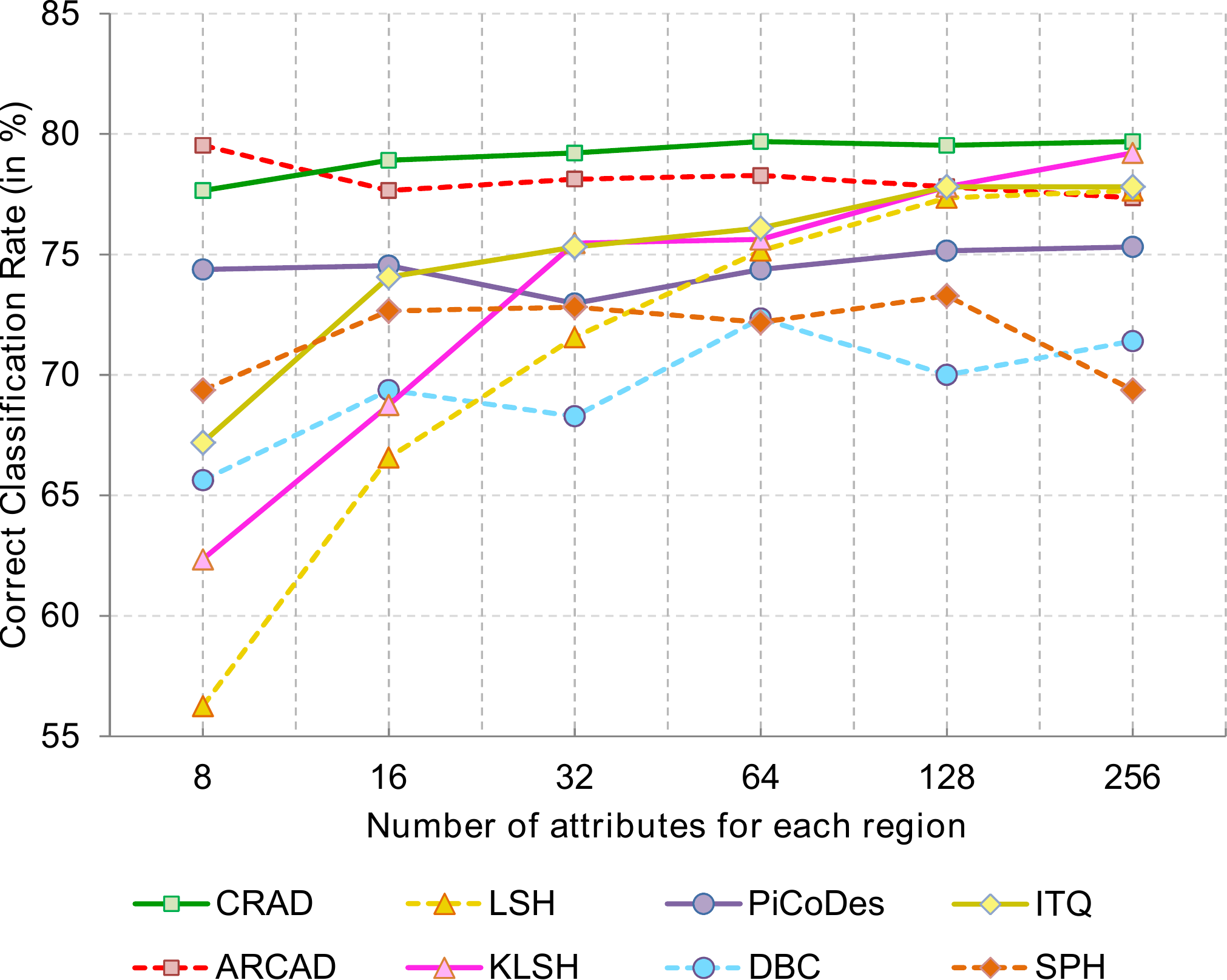}
  \caption {Performance comparison of the proposed approaches (ARCAD and CRAD) to the existing methods; LSH: Locality-Sensitive Hashing~\cite{charikar2002similarity}; KLSH: Kernelised LSH~\cite{kulis2009kernelized}; SPH: Spectral Hashing~\cite{weiss2008nips};ITQ: Iterative Quantization~\cite{gong2012}; DBC: Discriminative Binary Code~\cite{RastegariECCV2012}.}
  \label{fig:result_01}
\end{figure}

The results are presented in Fig.~\ref{fig:result_01}.
Both ARCAD and CRAD outperform all other methods even when small numbers of attributes are used.
This suggests that the proposed approaches are highly discriminative even with small code length.
The learning schemes successfully discover the essential cell attributes which are highly discriminative to form specimen image descriptor.
We note that both LSH and KLSH require longer code length to achieve similar performance to the proposed approaches.
This is consistent with the finding reported in previous works suggesting that LSH requires longer code length to achieve good performance~\cite{kulis2009kernelized}.
In our case having a small code length is advantageous since the experts require less time to name the discovered attributes.

It is noteworthy to mention that the proposed approaches outperform significantly PiCoDeS as well as DBC which were specifically designed to discover the discriminative attributes.
Both PiCoDeS and DBC consider attribute value in binary space.
Furthermore, the DBC learning scheme operates in the binary space which is intrinsically more complex.
In our case, we consider real attribute values (Refer Eqn.~\ref{eqn:h_j}).
In the light of this fact, we argue that presenting attribute 
value as a the real number is more expressive than a binary value (~\ie 0 or 1).

\begin{figure}[!tb] 
\centering
\includegraphics[width=0.7\columnwidth]{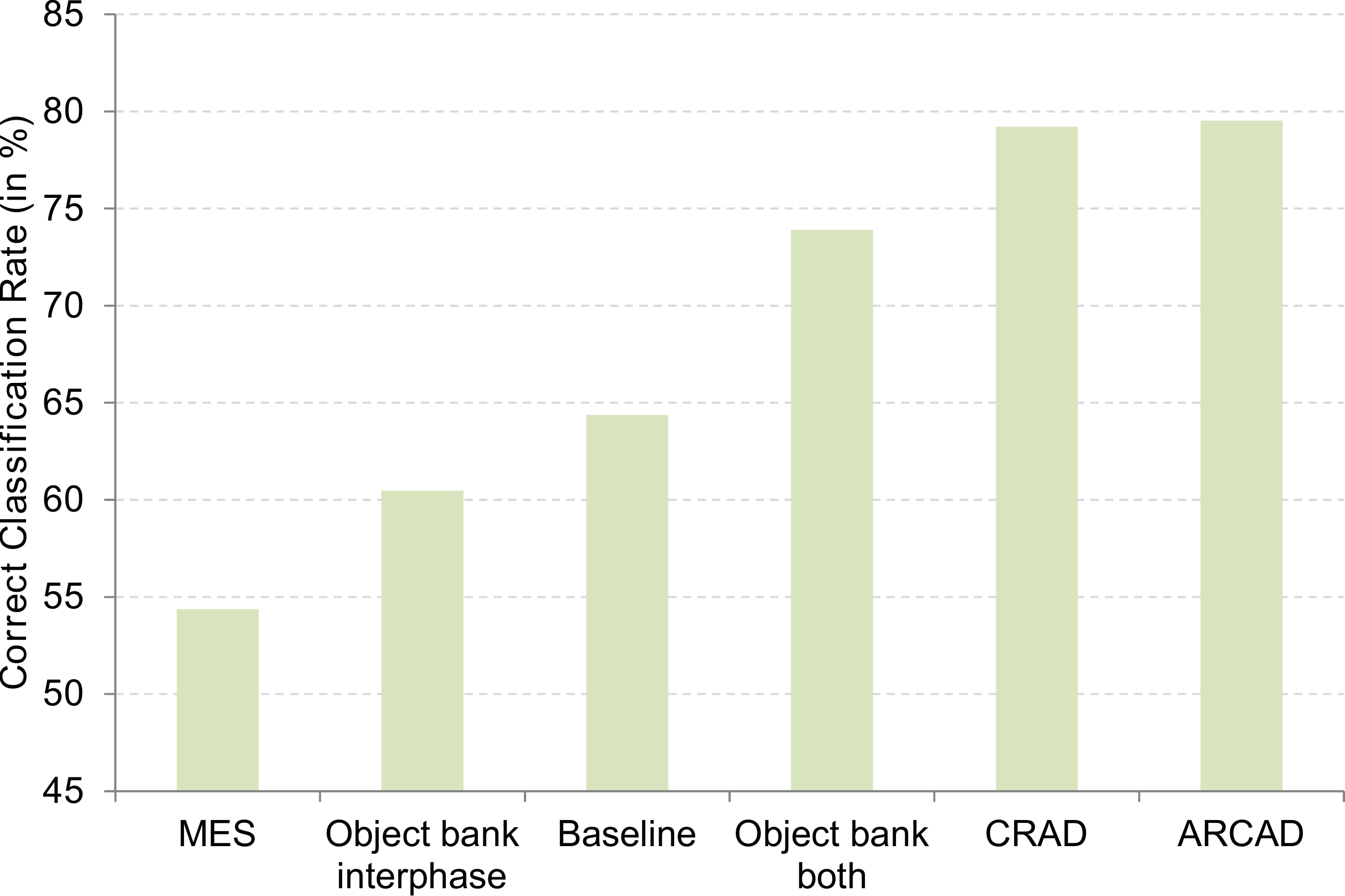}
  \caption {Performance comparison of the proposed approaches (ARCAD and CRAD) to the state of the arts; MES: Multiple Expert Systems~\cite{Soda2009}; Object bank proposed in~\cite{Li2013}.}
  \label{fig:result_02}
\end{figure}


In the second evaluation we contrasted the proposed approaches to existing methods for HEp-2 specimen image classification.
The most common approach for classifying specimen images is to use the dominant pattern of the interphase cells~\cite{foggia2013benchmarking,Wiliem2013,
Faraki2013,Yang2013}. 
Here, we call this \textit{baseline}.
Another approach, here denoted Multiple Expert System (MES), is to train individual classifier for each class (\ie one interphase cell classifier for each class) and use the classification reliability score to do weighted voting~\cite{Soda2009}.
We implemented both \textit{baseline} and the MES approach.

We also implemented the approach in~\cite{Li2013} which proposes the concept of Object Bank. 
Technically, we train k one-versus-all classifiers.
Given a cell image, we apply all the k classifiers and consider the k classification output scores as the object bank representation of the cell.
The object bank representation of a specimen image is obtained by averaging the {cell-level} object bank representation.
Here, we trained two sets of classifiers: (1) eight classifiers trained on interphase cells (\ie one classifier for each pattern class), denoted \textit{Object bank interphase}; (2) sixteen classifiers consisting of eight classifiers trained on the interphase cells and eight on the mitotic cells, denoted \textit{Object bank both}.

Fig.~\ref{fig:result_02} presents the evaluation results.
The proposed approaches significantly outperform all other methods.
We note that the MES has poor performance which contradicts what was reported in~\cite{Soda2009}.
Upon a closer look we found that some classes such as Mitotic Spindle do not have specific characteristics on their interphase cells.
Henceforth, it is difficult to train a reliable cell classifier rendering much lower reliability score.
In other hand, traditional voting (\ie the baseline) has much better performance as it only counts the vote and does not consider the classification reliability score.
Furthermore, we found that combining the information extracted from both interphase and mitotic cells is of importance.
This can be observed from the fact that there is a significant increase in Object bank performance when information from both cell types is used.


\vspace{-1ex}
\subsection{Describing ANA pattern class}

In this section we use the attributes discovered by the proposed approaches to generate a textual description of the eight ANA patterns.
To that end, we use attributes trained by CRAD as it gives the most consistent results in the previous evaluation.
Specifically, we opt to use 32 attributes extracted from inner and outer regions from both cell types.

\begin{figure}[tb]
\centering
\includegraphics[width=0.7\columnwidth]{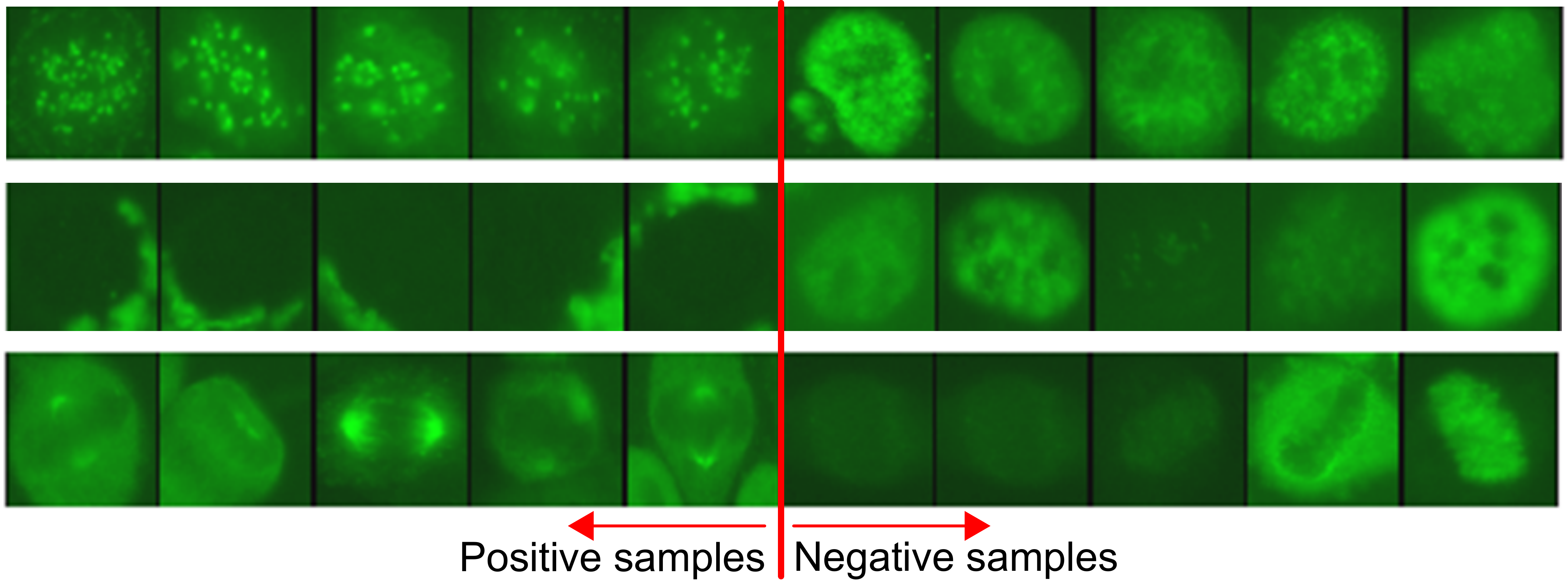}
  \caption {Example of successfully described attributes. from top row to bottom: \textit{parts of chromosome are stained}, \textit{golgi organelle is stained}, \textit{mitotic spindle staining}.}
  \label{fig:result_04}
\end{figure}

We first selected the most frequently appearing {cell-level} attributes from each pattern.
From the selected set, we further excluded the attributes which appear in at least more than four classes.
Finally, to name the cell attributes, we presented each cell attribute to the domain experts who were trained to read ANA by showing them both images classified as positive and negative by the attribute classifier.
We note that we presented the cell images in green colour which is similar to the colour of an ANA specimen under a fluorescent microscope.
Since the attributes are extracted from each cell region, we could ask more specific questions to the experts in relation to each region (\eg \textit{Please describe the property appearing at the cell boundary}).
The experts could opt not to name an attribute if they were not able to find any consistent property in the positive cell images. 
Fig.~\ref{fig:result_04} and~\ref{fig:examples} present some examples of cell attributes successfully identified by the experts.

Once the description for each class was generated, we let the experts indicate the correctness of each text description.
Fig.~\ref{fig:result_03} presents the generated description of each pattern.
Most patterns could be reasonably described with minor errors or omissions in the description.
The mitotic spindle pattern was perfectly described with no errors or omissions in the description.
On the other hand, despite this system being able to detect the important property of Golgi (\ie golgi organelle is stained), the system had more mistakes on Golgi than the other patterns. 
This is probably due to the fact that the Golgi pattern has only one prominent property.

\vspace{-2ex}

%% file: sec_conclusion.tex
\section{Main Findings}
\label{sec:conclusions}

The ANA test via Indirect Immunofluoresence protocol has been the gold standard to identifying Connective Tissue Diseases.
Unfortunately the protocol is subjective, time as well as labour intensive.
Despite the growing interest in this domain, prior works have primarily focused on classifying cell images extracted from specimen images.
In this work, we took a further step by addressing the specimen image classification problem.
To this end, we designed a specimen-level image descriptor to be: highly discriminative; short in descriptor length as well as semantically meaningful at cell level.
We achieved this goal by proposing two learning schemes which are based on the max-margin framework.
We later showed that under a certain condition, discovering such an image descriptor is equivalent to discovering discriminative image-level attributes.

We contrasted the proposed approaches to numerous hashing techniques as well as discriminative attribute learning approaches on a new HEp-2 cell dataset.
We found that the descriptors trained by the proposed schemes outperform the existing approaches with a much shorter code length.
The experiments also show that the proposed approaches outperform the recent approaches presented in~\cite{Soda2009,Li2013,Wiliem2013}.
Furthermore, we found that the descriptors can be used to provide a textual description of ANA patterns based on the cell attributes.

Despite the fact that the proposed approaches are able to discover semantically meaningful cell-level attributes, the constraints used to achieve this are not explicitly imposed in the proposed max-margin based training schemes.
As such, we will explore ways to make the constraints more explicit and study the effect of the constraints on the learned descriptor.

%% file: sec_acknowledgement.tex
\vspace{-2ex}
\section*{Acknowledgements}
\label{sec:Acknowledgements} 

This research was funded by Sullivan Nicolaides Pathology, Australia and the Australian Research Council (ARC) Linkage Projects Grant LP130100230. 

%% file: sec_appendix.tex
\appendix
\section*{Appendix}

\label{app:appendix}
\label{proof:ARCAD}
\noindent
We show the proof of Proposition.~\ref{prop:1} by deriving Eqn.~\ref{eqn:PiCoDes} from Eqn.~\ref{eqn:ARCAD}.
We first rewrite Eqn.~\ref{eqn:ARCAD}:

\begin{small}
\begin{equation}
\label{app:ARCAD}
\min_{\Mat{w}_{1 \dots K}, \Vec{b}_{1 \dots K}, \Mat{A}_{1 \dots J}} 
	\sum_{k=1}^K \left\{ \frac{1}{2} \| \Mat{w}_k \|^2 + \frac{\lambda}{N} \sum_{i=1}^N \ell \left[ y_{i,k} (\Vec{b}_k +
\sum^{J}_{j=1} \left( \frac{\Vec{w}_{k,j}}{N_{i,j}} \sum^{N_{i,j}}_{c=1} \Mat{A}^{\top}_j \Vec{x}_{i,j,c}  \right) \right] \right\}
\end{equation}
\end{small}

\noindent
This equation can be rewritten as:

\begin{small}
\begin{equation}
\min_{\Mat{w}_{1 \dots K}, \Vec{b}_{1 \dots K}, \Mat{A}_{1 \dots J}} 
	\sum_{k=1}^K \left\{ \frac{1}{2} \| \Mat{w}_k \|^2 + \frac{\lambda}{N} \sum_{i=1}^N \ell \left[ y_{i,k} (\Vec{b}_k +
\sum^{J}_{j=1} \left( \Vec{w}_{k,j} \Mat{A}^{\top}_j \left(  \frac{1}{N_{i,j}} \sum^{N_{i,j}}_{c=1}  \Vec{x}_{i,j,c} \right) \right) \right] \right\}
\end{equation}
\end{small}

\noindent
Let $\hat{\Vec{x}}_{i,j}$ be the average of cell-level descriptor extracted from $j$-th region 
$\hat{\Vec{x}}_{i,j} = \frac{1}{N_{i,j}} \sum^{N_{i,j}}_{c=1}  \Vec{x}_{i,j,c}$, the equation can be transformed into:


\begin{small}
\begin{equation}
\min_{\Mat{w}_{1 \dots K}, \Vec{b}_{1 \dots K}, \Mat{A}_{1 \dots J}} 
	\sum_{k=1}^K \left\{ \frac{1}{2} \| \Mat{w}_k \|^2 + \frac{\lambda}{N} \sum_{i=1}^N \ell \left[ y_{i,k} (\Vec{b}_k +
\sum^{J}_{j=1} \left( \Vec{w}_{k,j} \Mat{A}^{\top}_j \hat{\Vec{x}}_{i,j} \right) \right] \right\}
\end{equation}
\end{small}

\noindent
Finally, this can be rewritten into Eqn.~\ref{eqn:PiCoDes}, with $\Vec{u}_i = [\hat{\Vec{x}}_{i,j} \dots \hat{\Vec{x}}_{i,J}]$:

\begin{small}
\begin{equation}
\min_{\Mat{w}_{1 \dots K}, \Vec{b}_{1 \dots K}, \Mat{A}_{1 \dots J}} 
	\sum_{k=1}^K \left\{ \frac{1}{2} \| \Mat{w}_k \|^2 + \frac{\lambda}{N} \sum_{i=1}^N \ell \left[ y_{i,k} (\Vec{b}_k +
\sum^{P}_{p=1} \left( w_{k,p} \Mat{A}^{\top}_j \Vec{u_i} \right) \right] \right\}
\end{equation}
\end{small}

\begin{center}
\begin{figure*}[!tb] 
\includegraphics[width=\linewidth]{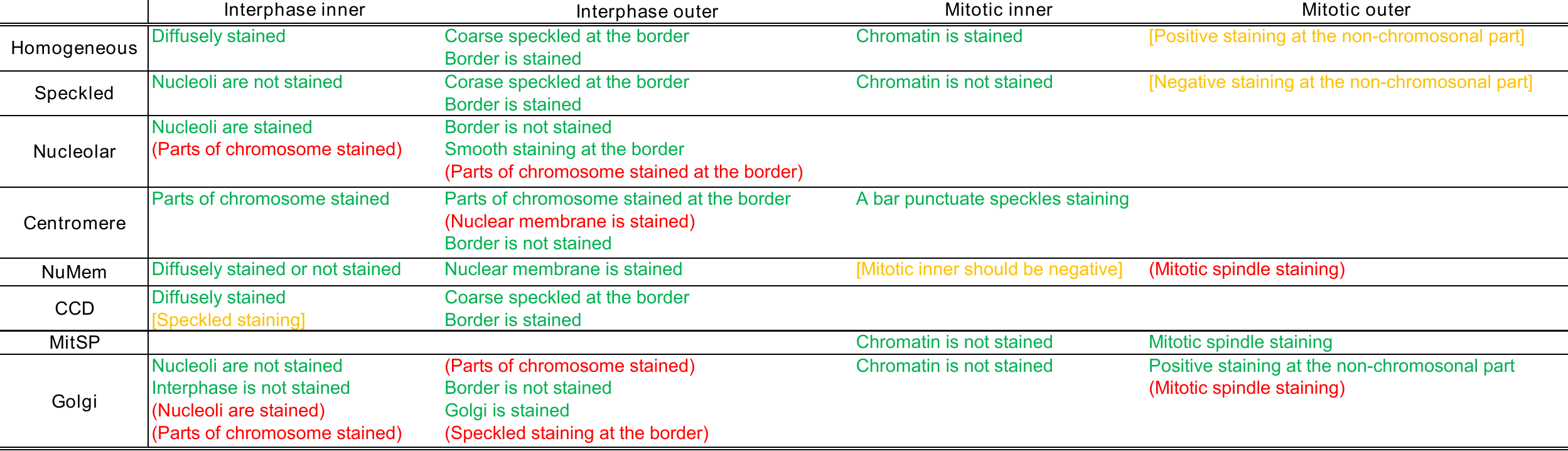}
  \caption {The generated text description for each pattern.
   The correct description is depicted in green colour and without bracket, whilst the text in square brackets and parentheses indicate the omitted and incorrect description, respectively.}
  \label{fig:result_03}
\end{figure*}
\end{center}